\documentclass{article}
\PassOptionsToPackage{numbers,sort&compress}{natbib}
\usepackage[dblblindworkshop,final]{neurips_infpriv_2026}
\usepackage{iftex}
\ifPDFTeX
  \usepackage[utf8]{inputenc}
\fi
\usepackage[T1]{fontenc}
\usepackage{amsmath,amssymb,booktabs,graphicx,microtype,xcolor}
\usepackage[hidelinks]{hyperref}
\usepackage{xurl,needspace,placeins}
\usepackage{etoolbox}
\makeatletter
\patchcmd{\@maketitle}{\vbox to \vsize}{\vbox}{}%
  {\PackageError{infpriv-title}{Title-box compatibility patch failed}{}}
\patchcmd{\@maketitle}{\@abstract}{}{}%
  {\PackageError{infpriv-title}{Abstract compatibility patch failed}{}}
\patchcmd{\@maketitle}{\vfill}{}{}%
  {\PackageError{infpriv-title}{Title-fill compatibility patch failed}{}}
\patchcmd{\ps@firstpage}{\begin{tabular}{c}}{\begin{tabular}{@{}c@{}}}{}%
  {\PackageError{infpriv-title}{Footer compatibility patch failed}{}}
\makeatother
\title{Auditing Post-Training Unlearning in Diffusion Language Models using Sparse Autoencoders}
\author{%
  Yunha Seo \\
  Department of Software \\
  Soongsil University \\
  \texttt{bobo9245@soongsil.ac.kr}
  \And
  Changhoon Kim \\
  Department of Software \\
  Soongsil University \\
  \texttt{kch@ssu.ac.kr}
}
\hypersetup{pdfauthor={Yunha Seo and Changhoon Kim},pdftitle={Auditing Post-Training Unlearning in Diffusion Language Models using Sparse Autoencoders}}
\newcommand{\E}{\mathbb E}
\newcommand{\KL}{\operatorname{KL}}
\newif\ifincludesupplement
\includesupplementtrue
\begin{document}
\maketitle

\begin{abstract}
Machine unlearning is an approach to privacy protection: it aims to remove the influence of designated data while preserving the model's behavior on information that should be retained. While a growing body of work investigates this objective in autoregressive large language models, machine unlearning remains underexplored in diffusion language models (DLMs). DLMs generate text through iterative denoising, exposing partially masked states whose internal representations change throughout the generation process. This raises a DLM-specific question: can selective forgetting improve by identifying forget-data-associated features and adapting their suppression across denoising stages? We conduct a controlled investigation of this question in LLaDA-8B-Instruct. To evaluate the unlearning of knowledge plausibly encoded during pre-training, we use WMDP as our forget-domain benchmark. Using top-$k$ sparse autoencoders and denoising-state replay, we compare static and stage-adaptive feature suppression. We find substantial shifts in forget-data-associated features across denoising stages, but tracking these dynamics alone does not reliably improve the forget-retain trade-off at the evaluated operating points. Together, our study provides an initial empirical foundation for developing and evaluating machine unlearning as a post-training privacy mechanism in DLMs.
\end{abstract}

\section{Introduction}
Removing designated information after model training is a practical concern for privacy-oriented AI. Machine unlearning addresses this goal by reducing the influence of forget data while preserving behavior on information that should be retained \citep{bourtoule2021machine}. When the original pre-training procedure cannot be rerun, behavioral audits can help assess whether a post-training intervention achieves selective removal. A decrease in target performance is informative, but its interpretation depends on whether unrelated capabilities remain available and whether the model can still produce meaningful outputs \citep{feng2025inconclusive}.

This assessment becomes more complex in diffusion language models (DLMs), which generate text by repeatedly updating a partially masked sequence \citep{nie2025llada}. Token probabilities and hidden representations evolve as additional context becomes visible. Recent work develops diffusion-specific unlearning objectives \citep{lee2026mdu}; a complementary question is whether representation-based interventions should change with the denoising stage. Sparse autoencoders (SAEs) support concept unlearning in text-to-image diffusion models \citep{cassano2026saemnesia,cywinski2025saeuron} and interpretability and steering in DLMs \citep{wang2026dlmscope}. Closely related, SAE--Guided Subspace Projection Unlearning (SSPU) uses forget/retain feature selection and QR-based decoder subspaces to guide parameter updates in autoregressive models \citep{wang2025sspu}. We inherit its activation-gap scoring, QR bases, and suppression through parameter updates, and adapt input construction and basis assignment to frozen diffusion states. Our novel contribution is a controlled audit of whether suppressing forget-associated SAE features, statically or by stage, can reduce designated knowledge while preserving unrelated capabilities.

Two assumptions require examination. First, a feature that activates more strongly on forget data may also identify a decoder direction important to retain data. Selection ranks sparsified encoder activations, whereas our intervention suppresses raw hidden-state projections onto decoder subspaces. Second, changes in selected features across stages do not by themselves show that matching suppression to those stages improves forgetting. An effective audit must therefore distinguish feature association, intervention selectivity, and the contribution of stage--basis alignment.

We study these questions in LLaDA-8B-Instruct using WMDP-Cyber \citep{li2024wmdp} and MMLU \citep{hendrycks2021mmlu}. Our controlled comparisons examine selected and random subspaces, pooled and stage-conditioned replay, and aligned and deliberately permuted stage assignments. WMDP provides a hazardous-knowledge proxy rather than a direct personal-data erasure test; our conclusions concern behavioral evidence for selective removal. We find that selected features are behaviorally consequential and change across denoising, yet neither property establishes an improved forgetting--utility trade-off. Output diagnostics further reveal that strong suppression can reduce measured accuracy while largely disabling generation.

\section{Post-Training Unlearning with SAE Subspaces}
\label{sec:protocol}
\paragraph{Setting.}
We study post-training removal when rerunning the original pre-training procedure is infeasible. WMDP-Cyber \citep{li2024wmdp} represents the designated knowledge to forget, and MMLU \citep{hendrycks2021mmlu} measures general task utility. We use WMDP-Cyber accuracy as a behavioral proxy for the suppression of the designated knowledge. This design tests removal of information already represented by the base model rather than information introduced through benchmark-specific fine-tuning.

\paragraph{Feature selection and controls.}
We use separately trained SAE checkpoints at layers 16, 24, and 30. For forget and retain corpora $D_F,D_R$, let $z_{\ell j}$ denote feature $j$'s activation at layer $\ell$. For each token, the checkpoint's TopK-32 SAE retains the 32 largest nonnegative activations. We then score every SAE feature by its mean squared activation on forget data minus that on retain data:
\begin{equation}
 s_{\ell j}=\E_{D_F}[z_{\ell j}^{2}]-\E_{D_R}[z_{\ell j}^{2}],
 \qquad Q_\ell=\operatorname{orth}\!\left(W_{\mathrm{dec},\ell}[\operatorname{top}_{64}(s_\ell)]^\top\right).
 \label{eq:selection}
\end{equation}
We call a feature forget-associated when it has a large positive activation gap. At each of layers 16, 24, and 30, Selected-SAE takes the 64 highest-scoring features, extracts their decoder directions, and orthonormalizes them with reduced QR to form the rank-64 basis $Q_\ell$. Thus, 32 denotes the number of SAE activations retained per token, whereas 64 denotes the number of decoder directions selected for suppression.

We use two random controls with different reference spaces. Gaussian samples 64 isotropic directions from the 4,096-dimensional hidden space at each layer and applies the same QR procedure. It tests the effect of suppressing an arbitrary subspace of the same rank. Random-SAE instead samples 64 decoder directions uniformly from the same SAE dictionary before QR. It controls for the geometry of the SAE dictionary while removing forget-based feature selection. Appendix~\ref{app:objectives} gives the exact encoder and sampling definitions.

\paragraph{Subspace Intervention.}
For each masked or replayed input $\tilde x$, we train the LoRA parameters $\theta$ \citep{hu2022lora} to reduce the projection of hidden states at token positions $V$ onto the selected subspace $Q_\ell^b$:
\begin{equation}
 \mathcal L(\theta)=\lambda\sum_\ell\frac{1}{|V|}\sum_{i\in V}
 \|h_{\theta,\ell,i}(\tilde x)Q_\ell^b\|_2^2
 +\mathcal L_{\mathrm{retain}}+10\mathcal L_{\mathrm{MCQ}}.
 \label{eq:projection}
\end{equation}
Here, $V$ contains the positions to which the projection penalty is applied, and $b$ identifies the basis used for the current input. The retention terms reconstruct a masked retain token and preserve the Base model's distribution over MMLU answer choices. All comparisons share initialization, retention data, and optimization settings.

\paragraph{Static and stage-dependent bases.}
The fixed-basis experiment selects one rank-64 basis per layer from clean text and applies it to every randomly masked training input. The replay experiments instead select features from early, middle, and late frozen-Base generation states. Pooled selection averages feature scores over the three stages and uses one shared basis per layer. Stage-conditioned selection constructs three bases per layer and applies the basis matched to the current denoising stage. Stage-permuted uses the same stage-specific bases but deliberately assigns them to the wrong stages, isolating the effect of stage-basis alignment.

We call an intervention selective when it lowers Cyber accuracy without a comparable reduction in retained utility. Feature association and intervention selectivity operate on different quantities: Equation~\eqref{eq:selection} ranks sparsified encoder activations, whereas Equation~\eqref{eq:projection} suppresses hidden-state energy in the span of the selected decoder directions. Appendix~\ref{app:geometry} gives an algebraic construction and measured examples of this distinction.

\paragraph{Evaluation.}
The fixed-basis experiment trains for 300 updates at $\lambda\in\{0.025,0.05,0.1,0.2,0.4\}$. The replay experiment compares pooled, stage-conditioned, stage-permuted, and Gaussian bases over 144 updates. Evaluation covers all 17,710 WMDP and MMLU questions; WMDP-Bio and WMDP-Chem measure collateral domain changes, and generated-answer tests audit output behavior. Appendix~\ref{app:objectives} provides the full protocol.

\section{Forget-Associated Directions Are Behaviorally Consequential}
\label{sec:results}

\begin{table}[t]
\centering
\caption{Accuracy (\%) after 300 updates at $\lambda=0.2$. Random-SAE results are reported as mean $\pm$ standard deviation over three independent basis draws, with the training seed held fixed.}
\label{tab:selection}
\small
\begin{tabular}{lrr}
\toprule
Basis & Cyber $\downarrow$ & MMLU $\uparrow$ \\
\midrule
Base & 43.03 & 64.61 \\
Gaussian & 32.36 & 62.19 \\
Random-SAE & 30.23 $\pm$ 0.75 & 61.69 $\pm$ 0.40 \\
Selected-SAE & 26.82 & 60.34 \\
\bottomrule
\end{tabular}
\end{table}

At $\lambda=0.2$, Selected-SAE reaches 26.82\% Cyber accuracy, compared with 32.36\% for Gaussian and $30.23\pm0.75$\% for Random-SAE (Table~\ref{tab:selection}). Relative to Gaussian, Selected-SAE reduces Cyber accuracy by an additional 5.54 pp, with a 95\% paired interval of $[-7.90,-3.12]$. At this coefficient, SAE selection therefore identifies directions whose suppression changes WMDP-Cyber accuracy more strongly than the Gaussian control.

The stronger target effect is coupled with a larger utility cost: Selected-SAE scores 60.34\% on MMLU, 1.85 pp below Gaussian with interval $[-2.52,-1.22]$. Across the full coefficient sweep, Selected-SAE does not dominate Gaussian in the Cyber--MMLU plane. Weak suppression produces similar Cyber accuracy with lower MMLU, while strong suppression drives both methods toward chance-level Cyber performance with larger utility loss for Selected-SAE (Figure~\ref{fig:selectiondynamics}a). Full coefficient and Random-SAE results appear in Appendix~\ref{app:completed}.

\paragraph{Output-level audit.}
At $\lambda=0.2$, Selected-SAE and Gaussian produce empty responses on 91\% and 90\% of the same 100 Cyber prompts. Generation collapse therefore accompanies the constrained-accuracy reduction under strong suppression, and lower multiple-choice accuracy alone does not certify information removal. Appendix~\ref{app:generation} reports valid-answer rates, paired outcomes, and answer-letter probability mass.

\begin{figure}[t]
\centering
\includegraphics[width=\linewidth]{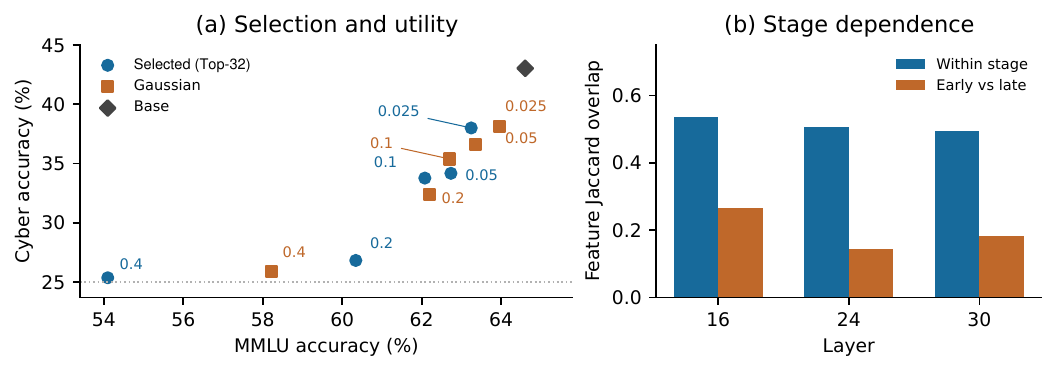}
\caption{\textbf{(a)} Cyber--MMLU trade-off across suppression coefficients; annotations give $\lambda$, and the dotted line marks 25\% chance accuracy. \textbf{(b)} Feature-selection stability under example resampling. Within-stage overlap exceeds early--late overlap at every evaluated layer.}
\label{fig:selectiondynamics}
\end{figure}

\section{Does Denoising-Aware Selection Improve Selectivity?}
\label{sec:temporal}
\paragraph{Forget-associated features shift across denoising.}
We select features from early, middle, and late frozen-Base generation states and bootstrap discovery examples to separate stage variation from selection noise. Within-stage feature Jaccard overlap ranges from 0.49 to 0.54, whereas early--late overlap ranges from 0.14 to 0.26 (Figure~\ref{fig:selectiondynamics}b). Decoder-subspace overlap follows the same ordering. The selected features and decoder subspaces therefore vary across the evaluated denoising states. Because scoring includes both masked and visible response positions, this variation reflects changing context and token visibility; it does not by itself establish that knowledge moves between features.

\paragraph{Stage-conditioned suppression.}
Pooled selection uses a single basis aggregated across stages, whereas stage conditioning applies a separate basis at each replay stage. A validation grid selects the operating point nearest to Base MMLU minus two percentage points. The selected methods lie within 0.51 pp of one another on validation, but none reaches the prespecified absolute target; we therefore report their observed trade-off rather than a utility-matched estimate.

\begin{table}[t]
\centering
\caption{Replay accuracy (\%) after 144 updates at validation-selected coefficients. Empty counts cover 16 Cyber and 16 MMLU generations.}
\label{tab:temporal}
\small
\begin{tabular}{lrrrr}
\toprule
Basis & $\lambda$ & Cyber $\downarrow$ & MMLU $\uparrow$ & Empty / 32 \\
\midrule
Base & -- & 43.08 & 64.70 & 0 \\
Pooled & 0.025 & 42.02 & 63.88 & 1 \\
Stage-conditioned & 0.025 & 40.92 & 63.53 & 0 \\
Gaussian & 0.1 & 41.72 & 63.39 & 2 \\
\bottomrule
\end{tabular}
\end{table}

Stage conditioning reaches 40.92\% Cyber and 63.53\% MMLU, compared with 42.02\% and 63.88\% for pooled replay (Table~\ref{tab:temporal}). It therefore couples an additional 1.11-pp Cyber reduction, $[-2.26,+0.05]$, with a 0.36-pp MMLU reduction, $[-0.69,-0.02]$. Relative to Gaussian, its Cyber difference is $-0.81$ pp, $[-2.26,+0.65]$. Stage conditioning changes target behavior but does not produce a utility-aware advantage at the evaluated operating points.

\paragraph{Alignment control.}
We permute early, middle, and late bases while preserving inputs, coefficients, and basis usage counts. Aligned suppression changes Cyber accuracy by $-0.70$ pp at $\lambda=0.025$, $[-1.76,+0.25]$, and by $-0.40$ pp at $\lambda=0.05$, $[-1.61,+0.86]$, relative to the permuted control. These comparisons do not establish a clear alignment benefit under the evaluated intervention; their intervals also do not establish that the effects are equivalent. Appendix~\ref{app:temporal_followup} reports the full validation grid, replay results, and alignment intervals.

\section{Additional Training-Seed and Strength Controls}
\label{sec:additional}
We extend the audit with three training seeds, four reference methods, and a Gaussian-matched control that equalizes initial hidden-gradient RMS. The SAE panel fixes the dictionary and bases, uses $\lambda=0.025$ and 144 updates, and varies unlearning seeds 42/43/44. Reference methods use 300 updates. Evaluation excludes the 25 Cyber generation-pilot questions, leaving 17,685 questions. Data splits and Gaussian coefficients differ from the primary replay study; Appendix~\ref{app:extensions} gives the protocol.

\begin{table}[t]
\centering\small
\caption{Additional training-seed panel: accuracy (\%), mean $\pm$ sample SD over unlearning seeds 42/43/44 with fixed SAE and bases. All five SAE conditions use $\lambda=0.025$ and 144 updates; Gaussian-matched adds Appendix~\ref{app:extensions}'s gains. Reference methods use 300 updates, own objectives.}
\label{tab:seeds}
\begin{tabular}{lrr}
\toprule
Method & Cyber $\downarrow$ & MMLU $\uparrow$ \\
\midrule
Base & 43.02 & 64.61 \\
Pooled & $41.17\pm0.64$ & $63.64\pm0.21$ \\
Stage-conditioned & $41.44\pm0.49$ & $63.52\pm0.23$ \\
Random-SAE & $42.52\pm0.21$ & $63.64\pm0.11$ \\
Gaussian & $41.10\pm0.12$ & $63.52\pm0.16$ \\
Gaussian-matched & $40.52\pm0.13$ & $63.34\pm0.05$ \\
\midrule
RMU \citep{li2024wmdp} & $43.10\pm0.15$ & $64.77\pm0.08$ \\
GA \citep{yao2024unlearning} & $42.93\pm0.06$ & $64.66\pm0.02$ \\
NPO \citep{zhang2024npo} & $42.83\pm0.23$ & $64.56\pm0.04$ \\
MDU \citep{lee2026mdu} & $42.56\pm0.13$ & $64.60\pm0.11$ \\
\bottomrule
\end{tabular}
\end{table}

\paragraph{Training-seed comparisons.}
Pooled reduces mean Cyber accuracy more than Random-SAE and Stage-conditioned, while Gaussian matches its reduction and Gaussian-matched exceeds it with more MMLU loss (Table~\ref{tab:seeds}). The selection benefit depends on the control. Reference methods remain within 0.46 pp of Base on Cyber under their reported settings. In the three-seed pilot, Pooled and Gaussian-matched have similar Cyber generation accuracy (20.00\% vs. 21.33\%), but Pooled is lower on MMLU (41.33\% vs. 52.00\%; Appendix~\ref{app:extendedgeneration}).

\begin{table}[t]
\centering\small
\caption{Post-hoc fresh layer-16 SAE (\%): one seed, 144 updates, learning rate $2\times10^{-5}$; utility is not matched. Cyber/MMLU use the 17,685-question panel; valid/empty use 25 Cyber generations.}
\label{tab:strengthaudit}
\begin{tabular}{lrrrrr}
\toprule
Method & $\lambda$ & Cyber $\downarrow$ & MMLU $\uparrow$ & Valid & Empty \\
\midrule
Base & -- & 43.02 & 64.61 & 92 & 0 \\
Pooled & 0.025 & 40.21 & 63.72 & 72 & 0 \\
Pooled & 0.1 & 38.23 & 63.17 & 40 & 28 \\
Random-SAE & 0.1 & 41.39 & 63.83 & 84 & 0 \\
Gaussian-matched & 0.1 & 42.61 & 63.66 & 88 & 0 \\
\bottomrule
\end{tabular}
\end{table}

\paragraph{Stronger suppression and generated answers.}
A post-hoc, single-seed exploration uses a fresh layer-16 top-32 SAE. Pooled $\lambda=0.025$ achieves lower Cyber accuracy than Gaussian-matched $0.1$ at comparable MMLU, with fewer valid Cyber answers (Table~\ref{tab:strengthaudit}). The 0.06 pp MMLU gap is not an advantage. Stronger Pooled suppression reduces Cyber further, while valid responses fall to 40\% and empty responses rise to 28\%. Appendix~\ref{app:extensions} details development selection and its bias.

\section{Conclusion}
We audit SAE-based post-training unlearning in LLaDA-8B-Instruct using separately trained SAEs. Selected directions reduce Cyber accuracy more strongly than random controls at the tested strong-suppression setting, accompanied by greater utility loss. Despite stage-dependent feature changes, the evaluated comparisons do not establish a clear benefit from stage conditioning or alignment. Geometry and output diagnostics further show why activation gaps and reduced accuracy are insufficient evidence for selective removal. Conditional on the tested intervention and training seeds, these findings motivate jointly auditing target behavior, retained capabilities, and output validity for privacy-oriented unlearning. The WMDP evaluation does not directly validate personal-data erasure or establish a privacy guarantee. Auditing suppression methods on forget sets that more directly instantiate personal information is a natural next step.

\label{maintextend}
\clearpage
\begingroup
\interlinepenalty=10000
\bibliographystyle{plainnat}
\bibliography{refs}
\endgroup
\ifincludesupplement
\clearpage
\appendix
\section{Experimental Setup}
\label{app:objectives}\label{app:data}\label{appendixstart}
This appendix describes the primary experiments. Protocols for the additional experiments are in Appendix~\ref{app:extensions}.
\paragraph{Feature definition.}
The SAE checkpoints were trained separately for this study at layers 16, 24, and 30. DLM-Scope is cited as related work, not as the source of these checkpoints. At each evaluated layer $\ell\in\{16,24,30\}$, the top-$k$ SAE is applied tokenwise to hidden states $h_{\ell i}$. It computes
\begin{equation}
 a(h)=\operatorname{ReLU}\!\left(((h-\mu)/\sigma)W_{\mathrm{enc}}+b_{\mathrm{enc}}\right),
 \qquad z(h)=\operatorname{TopK}_{32}(a(h)),
\end{equation}
using the normalization statistics stored in the SAE checkpoint. Here $\operatorname{TopK}_{32}$ retains the 32 largest nonnegative coordinates for each token and sets all others to zero. Feature scores are token means of forget-minus-retain squared top-32 activation. Reduced QR of the selected decoder atoms gives a rank-64 subspace at each layer. Gaussian samples 64 independent isotropic directions in the full 4,096-dimensional hidden space before QR. Random-SAE samples 64 distinct IDs uniformly from each 65,536-feature decoder dictionary and applies the same QR construction.

The checkpoint decoder bias is not used by the intervention: training suppresses raw hidden projections onto selected decoder atoms and does not reconstruct through the SAE.

\paragraph{Training data.}
Feature discovery in the fixed-basis study uses 540 Cyber texts from WMDP-Corpora \citep{li2024wmdp} and 540 general retain excerpts from WikiText-2-raw-v1 \citep{merity2017wikitext}. Optimization uses the first 300 forget and retain examples. The MCQ retention set contains 461 MMLU validation questions. Coefficient selection uses 1,029 additional validation questions across 57 subjects, excluding normalized exact-question matches to gradient-training and test questions.

\paragraph{Optimization.}
All methods start from adapter-free LLaDA-8B-Instruct \citep{nie2025llada} and identical initialized LoRA tensors \citep{hu2022lora}. LoRA covers available attention and MLP projection modules, with rank 32, alpha 64, zero dropout, and no bias updates. AdamW \citep{loshchilov2019adamw} uses learning rate $2\times10^{-5}$, betas $(0.9,0.999)$, epsilon $10^{-8}$, weight decay $0.01$, and a constant schedule without gradient clipping. Training and initialization seeds are 42. Fixed-basis training uses 300 updates; replay uses 144. Each comparison evaluates the final adapter.

For fixed-basis training, each update draws $p\sim\mathcal U(10^{-5},1)$ and masks forget tokens independently with probability $p$. Methods share corrupted inputs and data order. Retain reconstruction masks one deterministic midpoint token in a separate example. MCQ retention minimizes Base-to-student KL over the four answer choices at temperature one, weighted by ten. The projection term averages over positions and sums over layers, without initial-energy normalization.

\paragraph{Scoring and uncertainty.}
Evaluation contains 1,987 WMDP-Cyber, 1,273 WMDP-Bio, 408 WMDP-Chem, and 14,042 MMLU questions. MMLU is an unweighted macro average over 57 subjects. The prompt contains the raw question, choices, and answer prefix without a chat template. Left context is truncated at 512 tokens. Each answer letter is a single token, so the four candidates share one masked input and are scored by their full-vocabulary log probabilities at that position. All questions remain in the accuracy denominator.

Paired bootstrap intervals resample questions within each WMDP domain or separately within each MMLU subject. We use 1,000 replicates for fixed-basis comparisons and 2,000 for replay. Intervals condition on the evaluated models and questions; they exclude training-seed variation and do not adjust for model selection or multiple comparisons. Random-SAE's three basis draws measure basis variation with one shared training seed.

\section{Selection--Suppression Geometry}
\label{app:geometry}
\paragraph{Rank-one construction.}
Consider an SAE with identity normalization and zero encoder bias. Let $u\in\mathbb R^{1\times d}$ be a unit decoder row and use $u^\top$ as the encoder weight for coordinate $j$. Choose the remaining coordinates so that $j$ survives top-$k$ sparsification for $h_F=au$, where $0<a<b$. For $h_R=-bu$, its activation is zero after ReLU. Thus,
\begin{equation}
 z_j(h_F)^2-z_j(h_R)^2=a^2>0,\qquad
 \|h_F u^\top\|_2^2-\|h_R u^\top\|_2^2=a^2-b^2<0.
 \label{eq:gap}
\end{equation}
The feature has a positive forget-minus-retain activation gap, while suppression along its decoder direction assigns greater projection energy to retain data. The construction demonstrates that the selection score does not determine the ordering induced by the suppression objective.

\paragraph{Measured replay states.}
On 16 texts per corpus excluded from feature discovery, stage-selected top-32 activation-energy gaps summed over layers are approximately $+457$, $+364$, and $+342$ at early, middle, and late stages. The corresponding raw projection-energy gaps are $-905$, $-332$, and $+946$. The sign mismatch at the first two stages instantiates Equation~\eqref{eq:gap} in frozen-Base replay states. The activation and projection quantities use different numerical scales.

\FloatBarrier
\section{Full Fixed-Basis Comparisons}
\label{app:completed}\label{app:additional}

\begin{table}[!htbp]
\centering
\caption{Fixed-basis accuracy (\%) across suppression coefficients after 300 updates. Selected denotes Selected-SAE.}
\label{tab:frontier}
\small
\begin{tabular}{lrrrrr}
\toprule
Basis & $\lambda$ & Cyber & Bio & Chem & MMLU \\
\midrule
Selected & 0.025 & 38.00 & 64.57 & 46.32 & 63.25 \\
Gaussian & 0.025 & 38.10 & 64.73 & 47.79 & 63.96 \\
Selected & 0.05 & 34.17 & 63.79 & 43.63 & 62.74 \\
Gaussian & 0.05 & 36.64 & 64.26 & 46.57 & 63.36 \\
Selected & 0.1 & 33.77 & 64.18 & 45.59 & 62.08 \\
Gaussian & 0.1 & 35.38 & 64.18 & 46.08 & 62.71 \\
Selected & 0.2 & 26.82 & 60.96 & 40.44 & 60.34 \\
Gaussian & 0.2 & 32.36 & 63.94 & 46.08 & 62.19 \\
Selected & 0.4 & 25.36 & 50.75 & 34.31 & 54.10 \\
Gaussian & 0.4 & 25.87 & 56.40 & 38.24 & 58.22 \\
\bottomrule
\end{tabular}
\end{table}
\FloatBarrier

\paragraph{Utility and coefficient selection.}
The descriptive Selected $0.05$--Gaussian $0.10$ pair has an MMLU difference of $+0.027$ pp, with interval $[-0.412,+0.470]$. Its Cyber difference is $-1.21$ pp, with interval $[-2.57,+0.20]$. This pair was identified after test inspection and is descriptive rather than an independently selected utility-matched comparison.

A validation rule instead targets Base MMLU minus two percentage points and selects each method's nearest coefficient. Success requires both candidates within one point of the target and of each other. It selects Selected $0.2$ and Gaussian $0.05$; their validation gap of 1.49 pp fails the rule, and their test gap is 3.02 pp.

\begin{table}[!htbp]
\centering\small
\caption{MMLU validation accuracy (\%) on 1,029 questions. Bold entries are the validation-rule choices; the utility-matching criterion fails.}
\label{tab:grid}
\begin{tabular}{rrr}
\toprule
$\lambda$ & Selected-SAE & Gaussian \\
\midrule
0.025 & 64.287 & 63.439 \\
0.05 & 64.071 & \textbf{62.975} \\
0.1 & 64.084 & 63.019 \\
0.2 & \textbf{61.483} & 63.268 \\
0.4 & 53.910 & 58.969 \\
\bottomrule
\end{tabular}
\end{table}

\paragraph{Auxiliary controls.}
Retain-only removes the forget term while retaining the same initialization, retention objectives, optimizer, and 300-update budget. Its Cyber/MMLU accuracies of 43.08\%/64.53\% remain close to Base's 43.03\%/64.61\%, supporting attribution of larger changes to the suppression term.

\paragraph{Random decoder features.}
Random-SAE samples three independent decoder-feature sets and evaluates each at $\lambda\in\{0.05,0.2\}$. All other training conditions match the fixed-basis study.

\begin{table}[!htbp]
\centering
\caption{Uniform random SAE-feature selection: mean $\pm$ sample standard deviation across three basis draws with one shared training seed.}
\label{tab:randomsae}
\small
\begin{tabular}{rrrrr}
\toprule
$\lambda$ & Cyber & Bio & Chem & MMLU \\
\midrule
0.05 & 35.51 $\pm$ 1.74 & 64.65 $\pm$ 0.34 & 45.02 $\pm$ 1.48 & 63.00 $\pm$ 0.13 \\
0.2 & 30.23 $\pm$ 0.75 & 63.34 $\pm$ 0.12 & 44.36 $\pm$ 0.65 & 61.69 $\pm$ 0.40 \\
\bottomrule
\end{tabular}
\end{table}

\FloatBarrier
\section{Generation Diagnostics}
\label{app:generation}
Constrained answer-letter accuracy and generated answers measure different behaviors. The fixed-basis panel uses the same 100 Cyber questions for every model, a 16-position output span, greedy low-confidence remasking \citep{nie2025llada}, and zero guidance. Valid answers contain a single uppercase choice letter, optionally bracketed or preceded by a short answer prefix. Empty denotes no non-whitespace text after special-token removal.

\begin{table}[!htbp]
\centering
\caption{Generated accuracy, valid-answer rate, and empty-answer rate (\%) on the same 100 Cyber prompts.}
\label{tab:frontiergen}
\small
\begin{tabular}{lrrr}
\toprule
Model / coefficient & Gen. acc. & Valid & Empty \\
\midrule
Base & 26 & 78 & 0 \\
Selected 0.025 & 21 & 67 & 28 \\
Gaussian 0.025 & 16 & 57 & 34 \\
Selected 0.050 & 14 & 43 & 54 \\
Gaussian 0.050 & 20 & 48 & 45 \\
Selected 0.100 & 8 & 11 & 87 \\
Gaussian 0.100 & 18 & 46 & 49 \\
Selected 0.200 & 3 & 4 & 91 \\
Gaussian 0.200 & 7 & 10 & 90 \\
Selected 0.400 & 0 & 1 & 85 \\
Gaussian 0.400 & 1 & 1 & 61 \\
\bottomrule
\end{tabular}
\end{table}

\begin{table}[!htbp]
\centering\small
\caption{Paired diagnostics on the same 100 Cyber questions. Correct+empty is the joint event of a correct constrained decision and an empty generated answer.}
\label{tab:pairedoutputs}
\begin{tabular}{lrrrrr}
\toprule
Model & MCQ acc. & Gen. acc. & Empty & Correct+empty & \shortstack{Median\\letter mass} \\
\midrule
Base & 40 & 26 & 0 & 0 & 97.97 \\
Selected $0.20$ & 27 & 3 & 91 & 24 & 0.22 \\
Gaussian $0.20$ & 28 & 7 & 90 & 21 & 0.43 \\
\bottomrule
\end{tabular}
\end{table}

The answer-letter mass is $m_\theta=\sum_{a\in\{A,B,C,D\}}\exp(s_a)$, where $s_a$ is a full-vocabulary log probability at the masked position. It is not renormalized over the four choices. Under strong interventions, the low median mass shows that a constrained argmax selects among four low-probability letters. The paired records also contain correct constrained decisions with empty generated answers, separating the two evaluation modes.

\FloatBarrier
\section{Denoising-State Analysis and Replay}
\label{app:temporal}\label{app:temporal_followup}
\paragraph{State construction.}
Each corpus supplies 48 discovery texts and 16 additional texts excluded from feature discovery. The first 64 tokens form the prompt. Frozen Base generates 32 response tokens in one block using 16 greedy low-confidence denoising updates \citep{nie2025llada}. States before updates 0, 8, and 14 contain 32, 16, and 4 masked response positions. Clean discovery instead uses the original continuation. LLaDA receives no added time embedding.

For each stage, feature scoring averages over all response positions. The masked/visible composition changes across stages, so the overlap analysis measures selection changes across the complete replay states rather than at a fixed token-visibility pattern. Pooled selection averages the three stage-specific scores before selecting one rank-64 basis per layer. Stage conditioning uses the basis selected at the current stage, while Gaussian remains fixed. Each replay update uses one state; each forget example appears once at every stage over 144 updates. Suppression covers both masked and visible response positions. Frozen-Base replay does not follow trajectories generated by the updated model.

\paragraph{Selection stability.}
We bootstrap the 48 discovery examples independently within each corpus while retaining each sampled example's correspondence across stages. Within-stage overlap compares feature sets selected from two independent resamples at the same denoising stage; early--late overlap compares selections across the early and late stages. Feature Jaccard uses 200 replicates; subspace overlap uses the first 100.

\begin{table}[!htbp]
\centering
\caption{Example-resampling means. Subspace overlap is the mean squared principal-angle cosine.}
\label{tab:stability}
\small
\begin{tabular}{lrrrr}
\toprule
Layer & Jaccard: within & cross & Subspace: within & cross \\
\midrule
16 & 0.537 & 0.265 & 0.707 & 0.446 \\
24 & 0.507 & 0.144 & 0.675 & 0.291 \\
30 & 0.494 & 0.183 & 0.665 & 0.336 \\
\bottomrule
\end{tabular}
\end{table}

\paragraph{Comparison design.}
Pooled, stage-conditioned, and stage-permuted SAE bases are compared with Gaussian replay at $\lambda\in\{0.0125,0.025,0.05,0.1\}$. The permutation assigns bases $(14,0,8)$ to input stages $(0,8,14)$ while preserving each basis's usage count. Same-coefficient alignment comparisons therefore change only the input--basis pairing.

\paragraph{Validation selection.}
On 1,029 MMLU validation questions, each method selects the coefficient nearest to Base minus two percentage points. Base scores 64.27\%, giving a 62.27\% target. Success requires every selected value within one point of the target and their range at most one point. Their range is 0.51 pp, but each exceeds the target by more than one point, so the full matching criterion fails.

\begin{table}[!htbp]
\centering
\caption{MMLU validation macro accuracy (\%). Bold entries minimize distance to Base minus two points.}
\label{tab:temporalvalidation}
\small
\begin{tabular}{rrrrr}
\toprule
$\lambda$ & Pooled & Stage & Gaussian & Permuted \\
\midrule
0.0125 & 64.26 & 65.01 & 63.65 & 64.08 \\
0.025 & \textbf{63.89} & \textbf{63.49} & 63.86 & 63.88 \\
0.05 & 64.32 & 63.68 & 64.16 & 63.63 \\
0.1 & 64.00 & 63.72 & \textbf{63.38} & \textbf{63.34} \\
\bottomrule
\end{tabular}
\end{table}

\paragraph{Full evaluation and alignment.}
Table~\ref{tab:temporalfollowup} reports Base, each method's validation choice, and the additional conditions used for alignment comparisons. Inputs and suppression spans differ from the 300-update fixed-basis study, so effects are interpreted within replay.

\begin{table}[!htbp]
\centering
\caption{Full replay accuracy (\%) after 144 updates.}
\label{tab:temporalfollowup}
\small
\begin{tabular}{lrrrrr}
\toprule
Basis & $\lambda$ & Cyber & Bio & Chem & MMLU \\
\midrule
Base & -- & 43.08 & 65.67 & 45.83 & 64.70 \\
Pooled & 0.025 & 42.02 & 65.04 & 46.57 & 63.88 \\
Stage-conditioned & 0.025 & 40.92 & 64.81 & 46.57 & 63.53 \\
Stage-conditioned & 0.05 & 40.92 & 64.96 & 46.32 & 63.52 \\
Gaussian & 0.1 & 41.72 & 65.04 & 47.06 & 63.39 \\
Permuted & 0.025 & 41.62 & 64.73 & 46.32 & 63.67 \\
Permuted & 0.05 & 41.32 & 64.73 & 46.08 & 63.23 \\
Permuted & 0.1 & 41.62 & 65.28 & 46.32 & 63.22 \\
\bottomrule
\end{tabular}
\end{table}

\begin{table}[!htbp]
\centering
\caption{Stage-conditioned minus permuted differences (pp), with pointwise 95\% paired intervals.}
\label{tab:alignment}
\small
\begin{tabular}{lrr}
\toprule
$\lambda$ & Cyber: difference [interval] & MMLU: difference [interval] \\
\midrule
0.025 & -0.70 [-1.76, +0.25] & -0.14 [-0.46, +0.18] \\
0.05 & -0.40 [-1.61, +0.86] & +0.29 [-0.04, +0.62] \\
\bottomrule
\end{tabular}
\end{table}

Stage-conditioned minus Gaussian MMLU is $+0.14$ pp, $[-0.28,+0.53]$. The alignment comparisons change neither inputs nor basis usage counts, but their intervals include zero on both metrics.

\FloatBarrier
\section{Additional Experiment Protocols and Diagnostics}
\label{app:extensions}\label{app:extendedgeneration}
\paragraph{Training-seed panel and evaluation sets.}
The added panel in Section~\ref{sec:additional} uses unlearning seeds 42, 43, and 44 with fixed SAE dictionaries and bases. All five SAE conditions use base coefficient $\lambda=0.025$ and 144 updates. Basis/control seed 20261026 and decoding seed 42001 are fixed.

Per domain, disjoint corpus roles contain 300 training documents, 192 feature-discovery documents, 48 SAE-validation documents, and 60 hyperparameter-validation documents. Training truncates at 256 tokens. Replay scoring uses 48 feature-role documents; training replay draws separately from the training role. MCQ retention uses 461 MMLU questions; development uses 51. These settings differ from the primary protocols in Appendices~\ref{app:objectives}--\ref{app:temporal}.

We interpret comparisons within each panel. The primary validation-selected Gaussian uses $\lambda=0.1$, whereas the added seed panel uses $0.025$. Even at the shared $\lambda=0.025$, Stage-conditioned has lower Cyber accuracy than Pooled in Table~\ref{tab:temporal} but higher mean Cyber accuracy in Table~\ref{tab:seeds}. Empty outputs are also more frequent in the three-seed pilot of Table~\ref{tab:seedempty} than in Table~\ref{tab:temporal} (Pooled: 31/150 versus 1/32; Stage-conditioned: 50/150 versus 0/32). We do not attribute these between-panel differences to seed variation alone.

The added panels exclude the 25 Cyber generation-pilot questions, leaving 1,962 Cyber and 17,685 total. Each change uses its panel's Base. The primary fixed-basis and replay Base scores differ slightly (43.03/64.61 versus 43.08/64.70) because of device-dependent numerical differences; cross-device predictions disagree on 20 of 2,358 overlapping questions. Generation panels remain distinct: the primary fixed-basis study uses 100 Cyber questions, primary replay uses 16 Cyber and 16 MMLU questions, the added three-seed pilot uses 25 per domain, and strength selection uses 51 MMLU development questions.

\paragraph{Gaussian-matched control.}
For frozen replay states and orthonormal bases, define
\begin{equation}
 E_{b,\ell}=\E_h\|h_\ell Q_\ell^b\|_2^2,\qquad
 \lambda_{\mathrm{Gaussian},\ell}=\lambda\sqrt{\frac{E_{\mathrm{selected},\ell}}{E_{\mathrm{Gaussian},\ell}}}.
 \label{eq:gradientmatch}
\end{equation}
The squared-projection gradient is $2\lambda hQQ^\top$, with squared norm $4\lambda^2\|hQ\|_2^2$. The square-root ratio equalizes initial hidden-gradient RMS under common averaging. Layer-16/24/30 multipliers are 1.29/1.23/1.34; the fresh layer-16 SAE uses 1.60. This calibration does not match final utility.

\paragraph{Reference-method settings.}
RMU \citep{li2024wmdp}, GA \citep{yao2024unlearning}, NPO \citep{zhang2024npo}, and MDU \citep{lee2026mdu} use 300 updates. RMU uses steering-vector norm 20 and retain activation-MSE weight 100 at block 7, with LoRA restricted to blocks 5--7. GA and NPO use midpoint-token retain cross-entropy with weight one; NPO uses $\beta=0.1$ in the masked-CE surrogate. MDU uses its late-position forget objective and retain SFT term with weight one. These settings omit SAE's four-choice KL retention term. Their objectives, retention scope, and budgets differ from SAE replay; the reported points are not a common-budget tuning study.

\paragraph{MCQ retention and interpretation.}
For Base and student distributions normalized over the four answer letters, the loss in Equation~\eqref{eq:projection} is $\mathcal L_{\mathrm{MCQ}}=\E_q\KL(p_0^q\|p_\theta^q)$ at temperature one. We chose weight 10 empirically and held it fixed across SAE comparisons. Evaluation questions differ from retention questions but share MMLU's benchmark and answer format.

Four-choice normalization leaves the letters' total vocabulary probability mass unconstrained. This may contribute to the reduced Cyber letter mass in Appendix~\ref{app:generation}; the same questions' constrained accuracy also falls from 40\% to 27\% for Selected. MMLU letter-mass measurements and an absolute-mass retention control would test this hypothesis.

\paragraph{SAE configurations and provenance.}
In the added panels, hidden-state index $L$ means the output of zero-indexed transformer block $L-1$.  Records of their pretraining corpus and complete training history were not retained. Exact checkpoint/hook metadata for the primary experiments were not preserved with their reported results; the added panel's mapping is not assigned to them retroactively.

The strength panel trains a fresh layer-16 top-32 SAE for 4,000 updates, batch size 256, seed 42, and learning rate $10^{-3}$ using normalized reconstruction MSE. Inputs are masked independently with probability $0.5$ and truncated to 256 tokens. Training and validation activations use separate corpus roles. The reference unlearning point uses rank 64, $\lambda=0.025$, 144 updates, and learning rate $2\times10^{-5}$. The fresh single-layer and primary three-layer SAE configurations differ; strength effects are interpreted within the fresh-SAE panel.

\paragraph{Post-hoc strength exploration.}
We evaluate twelve selected combinations of $\lambda\in\{0.1,0.4\}$, 144/432 updates, learning rates $2\times10^{-5}$ and selected $6\times10^{-5}$ settings, including two variants excluding special tokens from suppression. The 432-update conditions reuse the same cached states. Development uses 60 forget and 60 retain documents and 51 MMLU questions. Selection requires MMLU loss at most 5 pp, retain masked-CE increase at most 10\%, empty rate at most 2\%, valid-rate decrease at most 5 pp, and forget masked-CE increase at least 0.1 nats relative to Base. Only Pooled $\lambda=0.1$, 144 updates, and learning rate $2\times10^{-5}$ satisfies every criterion.

The selected point reduces development forget projection energy from 473.10 to 64.83. Strength-panel results are single-seed, post-hoc observations on benchmarks seen earlier.

\paragraph{Generation and development selection.}
Table~\ref{tab:pilotgen} reports the same 25 Cyber questions, 16 output tokens, and 16 denoising steps. At Pooled $0.1$, 7 of 25 Cyber responses are empty, whereas none are empty on the 51 MMLU development questions. Because the empty-rate threshold was applied to those same development questions when selecting the condition, this contrast is not evidence of target-specific failure.

\begin{table}[t]
\centering\small
\caption{Fresh-SAE strength panel: generation on the same 25 Cyber pilot questions (\%). Validity follows the strict answer format; the auxiliary parser (Aux.) also extracts choices from explanations or choice text. Empty outputs contain no non-whitespace text after special-token removal.}
\label{tab:pilotgen}
\begin{tabular}{lrrrr}
\toprule
Method & Strict acc. & Aux. acc. & Valid & Empty \\
\midrule
Base & 36 & 40 & 92 & 0 \\
Pooled, $\lambda=0.025$ & 32 & 40 & 72 & 0 \\
Pooled, $\lambda=0.1$ & 12 & 24 & 40 & 28 \\
Random-SAE, $\lambda=0.1$ & 28 & 36 & 84 & 0 \\
Gaussian-matched, $\lambda=0.1$ & 28 & 32 & 88 & 0 \\
\bottomrule
\end{tabular}
\end{table}

Table~\ref{tab:seedempty} reports both domains of the separate three-seed pilot. Pooled and Gaussian-matched have similar Cyber generation accuracy, but Pooled has lower MMLU accuracy and more empty MMLU outputs (12.00\% versus 4.00\%). All five conditions produce more empty outputs on Cyber than on MMLU, including both random controls; Random-SAE has the highest empty rate on both domains. These are descriptive means on fixed small question panels.

\begin{table}[!htbp]
\centering\small
\caption{Additional training-seed panel: mean generation rates (\%) over unlearning seeds 42/43/44, with 25 fixed questions per domain and seed. Acc. is strict generated accuracy.}
\label{tab:seedempty}
\begin{tabular}{lrrrrrr}
\toprule
& \multicolumn{3}{c}{Cyber} & \multicolumn{3}{c}{MMLU} \\
\cmidrule(lr){2-4}\cmidrule(lr){5-7}
Method & Acc. & Valid & Empty & Acc. & Valid & Empty \\
\midrule
Pooled & 20.00 & 68.00 & 29.33 & 41.33 & 85.33 & 12.00 \\
Stage-conditioned & 12.00 & 44.00 & 54.67 & 42.67 & 84.00 & 12.00 \\
Random-SAE & 12.00 & 21.33 & 78.67 & 34.67 & 68.00 & 30.67 \\
Gaussian & 28.00 & 66.67 & 33.33 & 52.00 & 97.33 & 2.67 \\
Gaussian-matched & 21.33 & 50.67 & 49.33 & 52.00 & 96.00 & 4.00 \\
\bottomrule
\end{tabular}
\end{table}

\FloatBarrier

\fi
\end{document}